\documentclass[conference]{IEEEtran}
\usepackage{cite}
\usepackage{amsmath,amssymb,amsfonts}
\usepackage{algorithmic}
\usepackage{graphicx}
\usepackage{textcomp}
\usepackage{xcolor}
\usepackage{booktabs} 
\usepackage{multirow}

\def\BibTeX{{\rm B\kern-.05em{\sc i\kern-.025em b}\kern-.08em
    T\kern-.1667em\lower.7ex\hbox{E}\kern-.125emX}}

\begin{document}

\title{QuIC: A Quantum-Inspired Interaction Classifier for Revitalizing Shallow CNNs in Fine-Grained Recognition}

\author{
    \IEEEauthorblockN{Cheng Ying Wu}
    \IEEEauthorblockA{\textit{Undergraduate Program in Intelligent Computing and Big Data} \\
    \textit{Chung Yuan Christian University}\\
    Taoyuan City, Taiwan \\
    yingbow0115@gmail.com}
    \and
    \IEEEauthorblockN{Yen Jui Chang}
    \IEEEauthorblockA{\textit{Master Program in Intelligent Computing and Big Data} \\
    \textit{Quantum Information Center} \\
    \textit{Chung Yuan Christian University}\\
    Taoyuan City, Taiwan \\
    aceest@cycu.edu.tw}
}

\maketitle

\begin{abstract}

Deploying deep learning models for Fine-Grained Visual Classification (FGVC) on resource-constrained edge devices remains a significant challenge. While deep architectures achieve high accuracy on benchmarks like CUB-200-2011, their computational cost is often prohibitive. Conversely, shallow networks (e.g., AlexNet, VGG) offer efficiency but fail to distinguish visually similar sub-categories. This is because standard Global Average Pooling (GAP) heads capture only first-order statistics, missing the subtle high-order feature interactions required for FGVC. While Bilinear CNNs address this, they suffer from high feature dimensionality and instability during training. To bridge this gap, we propose the Quantum-inspired Interaction Classifier (QuIC). Drawing inspiration from quantum mechanics, QuIC models feature channels as interacting quantum states and captures second-order feature covariance via a learnable observable operator. Designed as a lightweight, plug-and-play module, QuIC supports stable, single-stage end-to-end training without exploding feature dimensions. Experimental results demonstrate that QuIC significantly revitalizes shallow backbones: it boosts the Top-1 accuracy of VGG16 by nearly \textbf{20\%} and outperforms state-of-the-art attention mechanisms (SE-Block) on ResNet18. Qualitative analysis, including t-SNE visualization, further confirms that QuIC resolves ambiguous cases by explicitly attending to fine-grained discriminative features and enforcing compact intra-class clustering.
\end{abstract}


\section{Introduction}

\IEEEPARstart{D}{eep} Convolutional Neural Networks (CNNs) have revolutionized general object recognition \cite{krizhevsky2012imagenet, simonyan2014very}. However, in the domain of \textbf{Fine-Grained Visual Classification (FGVC)}—discriminating between subordinate categories such as bird species, car models, or aircraft variants—significant challenges remain. Unlike generic classification tasks (e.g., distinguishing a cat from a dog) \cite{deng2009imagenet}, FGVC relies heavily on identifying subtle, localized discriminative features. For instance, distinguishing an \textit{American Crow} from a \textit{Common Raven} in the CUB-200-2011 dataset \cite{wah2011caltech} requires analyzing minute differences in beak shape and throat plumage texture, often amidst significant background clutter.

While deep architectures (e.g., ResNet-101 \cite{he2016deep}, DenseNet \cite{huang2017densely}) possess the capacity to handle these nuances, their massive computational cost and memory footprint make them impractical for ubiquitous deployment on resource-constrained edge devices and IoT nodes. Consequently, there is a critical need for high-performance \textbf{shallow architectures} (e.g., ResNet18 \cite{he2016deep}, VGG16 \cite{simonyan2014very}).

However, a fundamental limitation persists in these legacy or lightweight networks: their classification heads typically rely on Global Average Pooling (GAP) followed by a linear fully connected layer. This first-order aggregation treats feature channels as independent entities. In FGVC, however, visual identity is often defined by the \textbf{co-occurrence} or \textbf{interaction} of specific features (e.g., a specific beak shape co-occurring with a distinct wing pattern). Linear classifiers inherently fail to capture these high-order correlations. While Bilinear CNNs (B-CNN) \cite{lin2015bilinear} address this by computing the outer product of feature maps, they suffer from dimensionality explosion (e.g., creating $262$k dimensional vectors) and typically require complex, multi-stage training pipelines to converge, rendering them prohibitive for real-time applications.

To bridge this gap, we draw inspiration from \textbf{Quantum Mechanics}. In quantum theory, the state of a complex system is often defined not just by individual components but by their \textbf{entanglement}—non-local interactions where the state of one particle is intrinsically linked to another. We hypothesize that feature channels in a neural network can be modeled as "quantum states," and their interaction holds the key to fine-grained discrimination. Based on this, we propose the \textbf{Quantum-inspired Interaction Classifier (QuIC)}. QuIC simulates the physical measurement process $\langle \psi | \mathcal{M} | \psi \rangle$, introducing a learnable interaction matrix that explicitly captures second-order correlations between features. Crucially, QuIC achieves this without the parameter explosion of traditional bilinear methods.

Our contributions are summarized as follows:
\begin{enumerate}
    \item \textbf{Plug-and-Play Covariance Module:} We propose QuIC, a lightweight, quantum-inspired module designed to capture second-order feature entanglement. Unlike B-CNNs \cite{lin2015bilinear} which often require complex two-stage training or high-dimensional feature maps, QuIC is fully differentiable and supports stable, \textbf{single-stage end-to-end training}. It can be easily inserted into standard backbones (e.g., AlexNet \cite{krizhevsky2012imagenet}, ResNet \cite{he2016deep}) as a drop-in replacement for the standard FC head.
    
    \item \textbf{Universal Performance Boost:} We demonstrate that QuIC consistently revitalizes various shallow backbones. On the challenging \textbf{CUB-200-2011} dataset \cite{wah2011caltech}, QuIC boosts the accuracy of AlexNet, VGG16 \cite{simonyan2014very}, GoogLeNet \cite{szegedy2015going}, and ResNet18 \cite{he2016deep}, proving its robustness and generalization capability across different architectures.
    
    \item \textbf{Interpretability and Precision:} Qualitative analysis confirms that QuIC's performance gains stem from its ability to model fine-grained interactions. Visualization results (Grad-CAM \cite{selvaraju2017grad} and t-SNE \cite{maaten2008visualizing}) show that QuIC successfully localizes minute discriminative parts and enforces compact feature clustering, providing a transparent explanation for its "Exclusive Wins" over attention-based baselines.
\end{enumerate}

\section{Related Work}

\subsection{Fine-Grained Visual Classification (FGVC)}
Fine-Grained Visual Classification aims to distinguish subordinate categories with subtle inter-class differences, such as bird species, car models, or aircraft variants. The field relies heavily on standard benchmarks like the \textbf{CUB-200-2011} dataset \cite{wah2011caltech}, which poses significant challenges due to high intra-class variance and low inter-class variance.

Early approaches relied heavily on part-based models that explicitly localized discriminative regions using pose normalization techniques. However, these methods often require expensive annotations (bounding boxes or keypoints). Recent trends have shifted towards learning discriminative representations directly from raw images using deep CNNs trained with only image-level labels. Architectures such as VGG \cite{simonyan2014very} and ResNet \cite{he2016deep} have become standard backbones for feature extraction, though they often require specialized heads to capture fine-grained details effectively.

\subsection{Bilinear Pooling and Feature Interaction}
To capture the subtle feature correlations essential for FGVC, Lin et al. proposed \textbf{Bilinear CNNs (B-CNN)} \cite{lin2015bilinear}. By computing the outer product of two feature maps, B-CNNs explicitly model the second-order interactions (co-occurrence) of features, significantly improving performance on texture-heavy tasks. However, standard B-CNNs suffer from two major limitations:
\begin{enumerate}
    \item \textbf{Dimensionality Explosion:} The resulting bilinear feature vector has a dimension of $C \times C$. For a standard feature dimension of $C=512$, this results in over 260k parameters, imposing a heavy memory burden.
    \item \textbf{Training Complexity:} B-CNNs often require a cumbersome two-stage training procedure or specific element-wise power normalizations to converge.
\end{enumerate}

Subsequent works like Compact Bilinear Pooling (CBP) \cite{gao2016compact} attempted to address the dimensionality issue using tensor sketching techniques (e.g., Count Sketch) to approximate the bilinear feature map. While effective at reducing dimensions, these approximations often introduce training instability or require complex frequency-domain computations. Unlike these methods, our proposed \textbf{QuIC} captures second-order interactions directly within the classification head without explicitly generating high-dimensional descriptors, supporting stable end-to-end training.

\subsection{Attention Mechanisms}
Attention mechanisms have become a dominant paradigm for enhancing feature representation. Notably, Squeeze-and-Excitation (SE) networks \cite{hu2018squeeze} adaptively recalibrate channel-wise feature responses by modeling inter-dependencies between channels.

While SE blocks successfully capture global context information, they primarily perform \textbf{first-order} feature reweighting (i.e., scaling importance). They do not explicitly model the specific "entanglement" or second-order feature interactions between pairs of features, which limits their effectiveness in distinguishing highly similar species where the \textit{combination} of features (e.g., a specific beak shape AND wing texture) is more discriminative than the presence of a single feature alone. QuIC overcomes this by mathematically modeling these pairwise interactions via a quantum-inspired observable operator.

\section{Methodology}

In this section, we formally introduce the \textbf{Quantum-inspired Interaction Classifier (QuIC)}. We first bridge the gap between classical feature extraction and quantum measurements, then detail the architectural implementation.

\subsection{Theoretical Formulation: Feature Vectors as Quantum States}

In a standard CNN pipeline, the backbone maps an input image to a feature tensor. After Global Average Pooling (GAP), this is reduced to a latent feature vector $\mathbf{z} \in \mathbb{R}^C$.
A classical linear classifier computes the logit $y_k$ for class $k$ as the inner product:
\begin{equation}
    y_k = \mathbf{w}_k^T \mathbf{z} + b_k
\end{equation}
This formulation implicitly assumes that features contribute independently to the class score. However, in FGVC, visual identity is often determined by the \textit{interaction} of features (e.g., specific beak shapes co-occurring with throat textures).

To model these high-order interactions, we draw an analogy to quantum mechanics. We propose treating the normalized feature vector $\mathbf{z}$ as a representation of a \textbf{quantum state vector} $|\psi\rangle$ in a Hilbert space. In quantum theory, the value of a physical quantity (observable) is determined by the expectation value of an \textbf{Observable Operator} $\mathcal{M}$:
\begin{equation}
    E = \langle \psi | \mathcal{M} | \psi \rangle = \mathbf{z}^T \mathcal{M} \mathbf{z}
    \label{eq:quantum_expectation}
\end{equation}
Here, the operator $\mathcal{M}$ encodes the measurement interactions. While $\mathcal{M}$ is a Hermitian matrix in complex quantum mechanics, in our real-valued neural network implementation, we model it as a \textbf{Symmetric Matrix}. The off-diagonal elements $\mathcal{M}_{ij}$ capture the \textbf{``entanglement''} (or second-order interactions) between the $i$-th and $j$-th feature channels.

\subsection{The QuIC Architecture}

Based on the formulation above, the QuIC module is designed as a plug-and-play head. It consists of two parallel pathways that are fused to produce the final class score:

\subsubsection{Linear Path (Superposition)}
This path preserves the first-order information, identical to a standard classifier:
\begin{equation}
    \mathcal{F}_{\text{linear}}(\mathbf{z})_k = \mathbf{w}_k^T \mathbf{z}
\end{equation}

\subsubsection{Entanglement Path (Interaction)}
To capture second-order interactions, QuIC learns a class-specific \textbf{Interaction Matrix} $\mathbf{M}_k \in \mathbb{R}^{C \times C}$ for each class $k$. The score contribution is computed via the quadratic form:
\begin{equation}
    \mathcal{F}_{\text{entangle}}(\mathbf{z})_k = \mathbf{z}^T \mathbf{M}_k \mathbf{z}
\end{equation}
This term explicitly models the second-order interactions between features. To ensure the operator behaves as a valid observable and to reduce parameter redundancy, $\mathbf{M}_k$ is constrained to be symmetric.

\subsubsection{Unified Output}
The final logit for class $k$ is the fusion of both the superposition (linear) and entanglement (interaction) paths:
\begin{equation}
    y_k = \text{BN}\left( \mathbf{w}_k^T \mathbf{z} + \mathbf{z}^T \mathbf{M}_k \mathbf{z} \right) + b_k
\end{equation}
where $\text{BN}$ denotes a Batch Normalization layer applied to the fused logits to stabilize training dynamics.

\subsection{Efficiency Analysis}

A key advantage of QuIC is its computational efficiency compared to traditional Bilinear CNNs (B-CNN). 
\begin{itemize}
    \item \textbf{B-CNN Approach:} Explicitly computes the outer product $\mathbf{x} = \mathbf{z}\mathbf{z}^T$, resulting in a high-dimensional feature descriptor of size $C^2$ (e.g., $512^2 \approx 262$k dimensions). This causes a memory bottleneck that often prevents end-to-end fine-tuning.
    \item \textbf{QuIC Approach:} Computes the scalar interaction score $\mathbf{z}^T \mathbf{M}_k \mathbf{z}$ directly via efficient bilinear transformations, avoiding the creation of massive intermediate tensors.
\end{itemize}
Consequently, QuIC is lightweight in terms of activation memory and supports stable, single-stage end-to-end training on standard GPUs.

\section{Experiments}

\subsection{Dataset and Implementation Details}
We evaluated our proposed method on the \textbf{CUB-200-2011} dataset \cite{wah2011caltech}, a standard benchmark for Fine-Grained Visual Classification containing 11,788 images of 200 bird species.
Input images were resized to \textbf{$448 \times 448$} resolution to capture sufficient fine-grained detail. All models were trained using Stochastic Gradient Descent (SGD) with a momentum of 0.9, weight decay of $1e-4$, and a batch size of 16 for 50 epochs. The learning rate was initialized at 0.001 and decayed by a factor of 0.1 every 15 epochs.

\subsection{Quantitative Results}
We evaluated QuIC across four backbone architectures: \textbf{AlexNet}, \textbf{VGG16}, \textbf{GoogLeNet}, and \textbf{ResNet18}. To validate the effectiveness of our module, we compared QuIC against three baselines:
\begin{itemize}
    \item \textbf{FC (Baseline):} Standard fully connected head on flattened features.
    \item \textbf{GAP:} Global Average Pooling followed by a linear classifier (standard for modern CNNs).
    \item \textbf{SE-Block:} Squeeze-and-Excitation attention module \cite{hu2018squeeze} added before the classifier.
\end{itemize}

\begin{table}[htbp]
\caption{Top-1 Accuracy Comparison on CUB-200-2011 ($448 \times 448$)}
\centering
\begin{tabular}{lcccc}
\toprule
\textbf{Backbone} & \textbf{FC} & \textbf{GAP} & \textbf{SE} & \textbf{QuIC (Ours)} \\
\midrule
AlexNet     & 52.4\% & 58.4\% & 57.9\% & \textbf{62.2\%} \\
VGG16       & 61.3\% & 77.0\% & 76.6\% & \textbf{81.0\%} \\
GoogLeNet   & 76.7\% & 78.4\% & 75.6\% & \textbf{80.3\%} \\
ResNet18    & 63.5\% & 78.8\% & 77.7\% & \textbf{80.5\%} \\
\bottomrule
\end{tabular}
\label{tab:main_results}
\end{table}

Table \ref{tab:main_results} summarizes the Top-1 classification accuracy. The results reveal two key insights:
\begin{itemize}
    \item \textbf{Universal Improvement:} QuIC consistently outperforms the FC and GAP baselines across all backbones. Notably, it provides a massive boost to VGG16 (+19.7\% over FC), effectively "revitalizing" this legacy architecture.
    \item \textbf{Superiority over Attention:} On ResNet18, QuIC achieves a Top-1 accuracy of \textbf{80.5\%}, surpassing the SE block (77.7\%). This confirms that explicitly modeling second-order interactions is more effective for fine-grained tasks than first-order channel reweighting.
\end{itemize}

\begin{figure}[htbp]
\centerline{\includegraphics[width=\columnwidth]{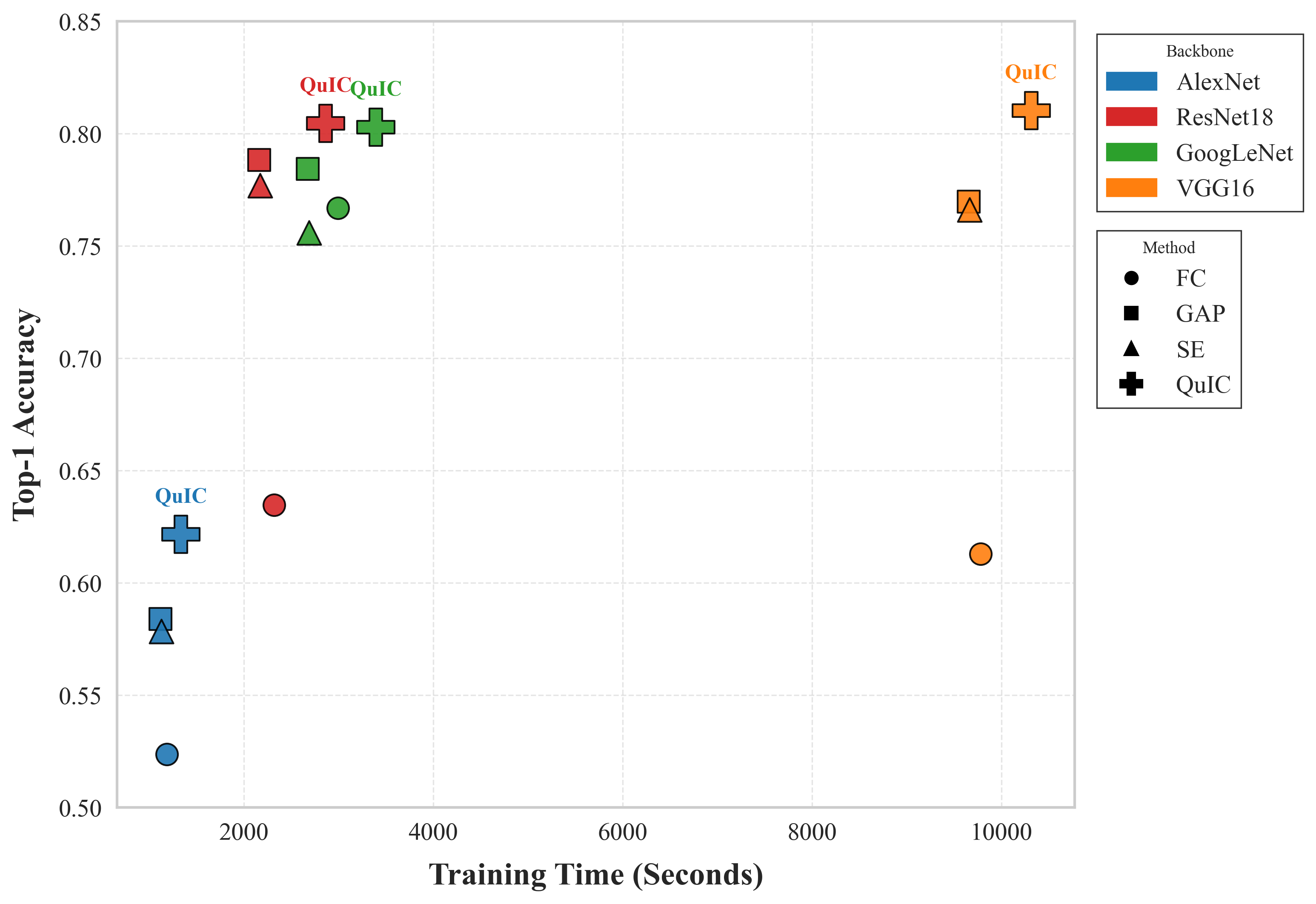}}
\caption{Efficiency Analysis: Accuracy vs. Training Time. QuIC achieves the best trade-off, offering superior accuracy with minimal computational overhead compared to attention-based baselines.}
\label{fig:efficiency}
\end{figure}

\subsection{Efficiency Analysis}
We further investigated whether the performance gains come at the cost of computational efficiency. Fig. \ref{fig:efficiency} illustrates the trade-off between training time (x-axis) and Top-1 accuracy (y-axis).
QuIC (marked with '+' achieves the highest accuracy cluster in the top-right corner. Crucially, its training cost is comparable to the lightweight GAP and SE baselines, confirming that QuIC avoids the heavy computational burden associated with traditional bilinear pooling.

\begin{figure}[htpb]
\centerline{\includegraphics[width=\columnwidth]{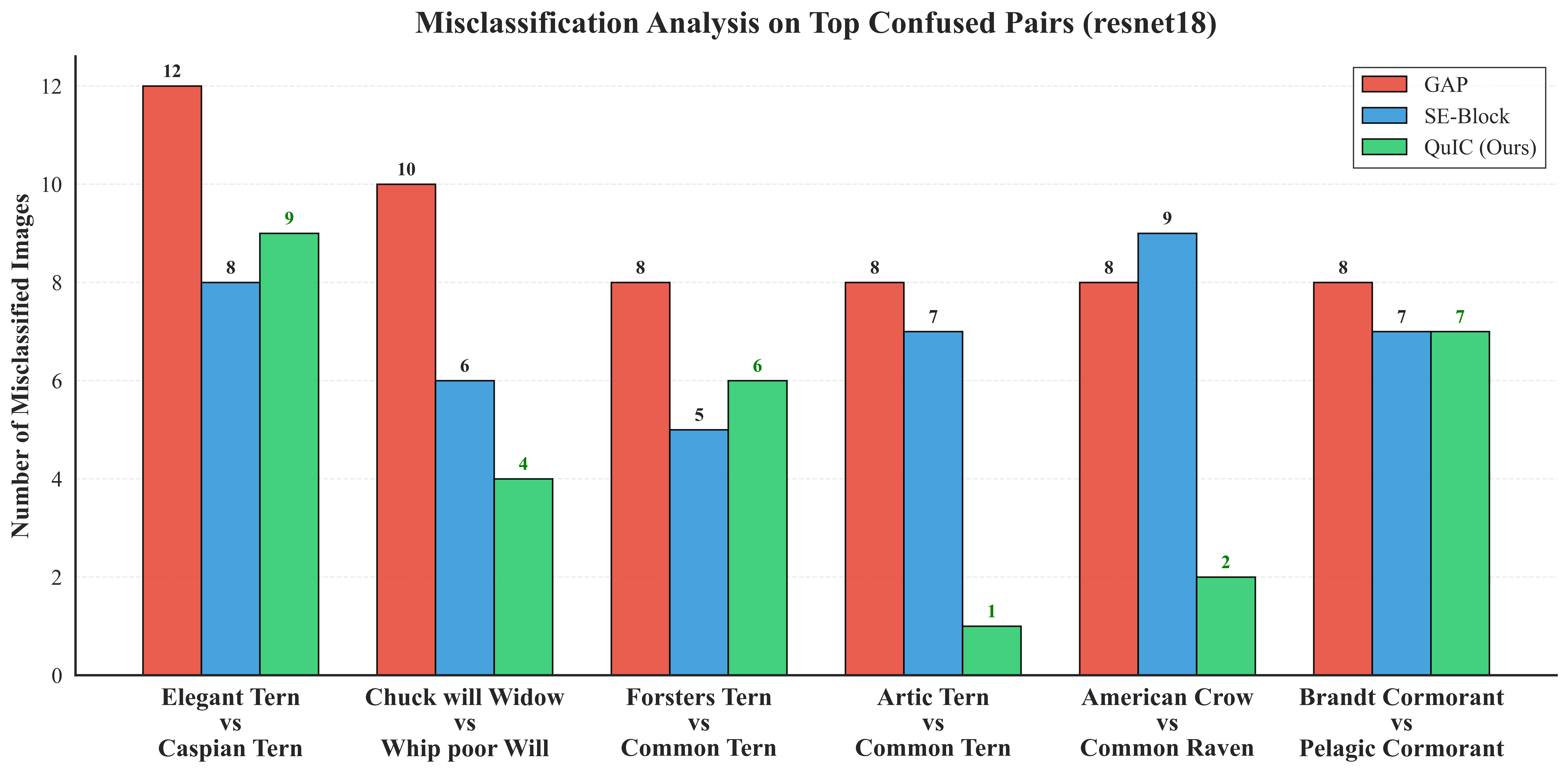}}
\caption{Comparison of misclassification counts for the top confused pairs using the ResNet18 backbone. QuIC (Green) demonstrates superior discriminative ability on highly similar species, significantly reducing errors on challenging pairs such as "American Crow vs. Common Raven", "Arctic Tern vs. Common Tern", and "Chuck-will's-widow vs. Common Poorwill" compared to GAP (Red) and SE (Blue) baselines.}
\label{fig:confusion}
\end{figure}

\begin{figure*}[t!] 
    \centering 
    \includegraphics[width=\textwidth, height=0.5\textheight, keepaspectratio]{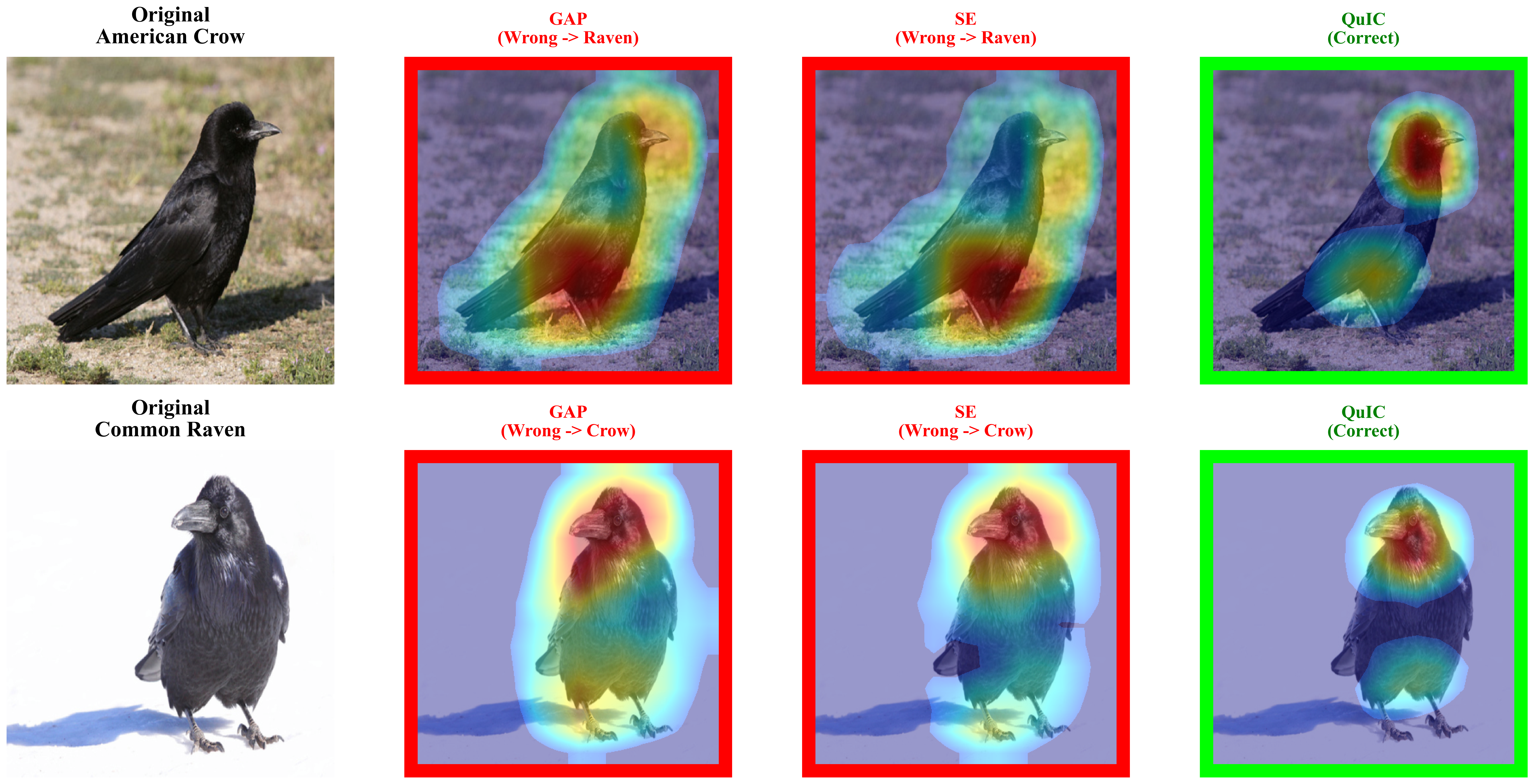}
    \caption{Visualization of "Exclusive Wins" for the (Top) American Crow and (Bottom) Common Raven. While baselines misclassify these samples (e.g., mistaking the Crow for a Cowbird and the Raven for a Crow) due to diffuse attention, QuIC successfully localizes fine-grained discriminative parts—specifically the beak shape and throat hackles—leading to correct predictions.}
    \label{fig:gradcam} 
\end{figure*}

\section{Qualitative Analysis}

To investigate the source of QuIC's performance gains, we conducted a detailed qualitative analysis, specifically focusing on its ability to distinguish challenging pairs and the structure of the feature space it learns.

\subsection{Confusion Analysis on Challenging Pairs}
The distinction between visually similar species pairs proved particularly challenging for baseline models. For instance, the \textit{American Crow} and \textit{Common Raven} share nearly identical all-black plumage, causing the GAP and SE models to misclassify 8 and 9 images, respectively, as illustrated in Fig. \ref{fig:confusion}. In contrast, QuIC significantly reduced this confusion, resulting in only 2 misclassified samples. This discriminative advantage extends to other challenging families: for the \textit{Arctic Tern} vs. \textit{Common Tern} pair, QuIC reduced misclassifications to just 1 (compared to 7--8 for baselines), and it similarly outperformed baselines on the cryptic \textit{Chuck-will's-widow} vs. \textit{Common Poorwill} pair. These results suggest that explicit second-order interaction modeling is essential for resolving the fine-grained structural ambiguities that confound standard first-order networks.

\subsection{Visualizing the Quantum Advantage}
We visualized class activation maps using Grad-CAM \cite{selvaraju2017grad}. Fig. \ref{fig:gradcam} presents \textbf{``Exclusive Win'' cases} where QuIC correctly classifies \textbf{an \textit{American Crow} and a \textit{Common Raven}} that other models missed.
\begin{itemize}
    \item \textbf{Baseline Failure:} GAP and SE models exhibit diffuse activation maps. They primarily focus on the global black shape or background context, which is insufficient for distinguishing the two species.
    \item \textbf{QuIC Success:} QuIC produces a highly localized map focusing on the \textbf{throat} and \textbf{beak} areas. Biologically, Common Ravens are distinguished from Crows by their "hackles" (shaggy throat feathers) and heavier beaks. QuIC successfully attends to these minute discriminative features.
\end{itemize}

\begin{figure}[t]
    \centering
    \includegraphics[width=\columnwidth]{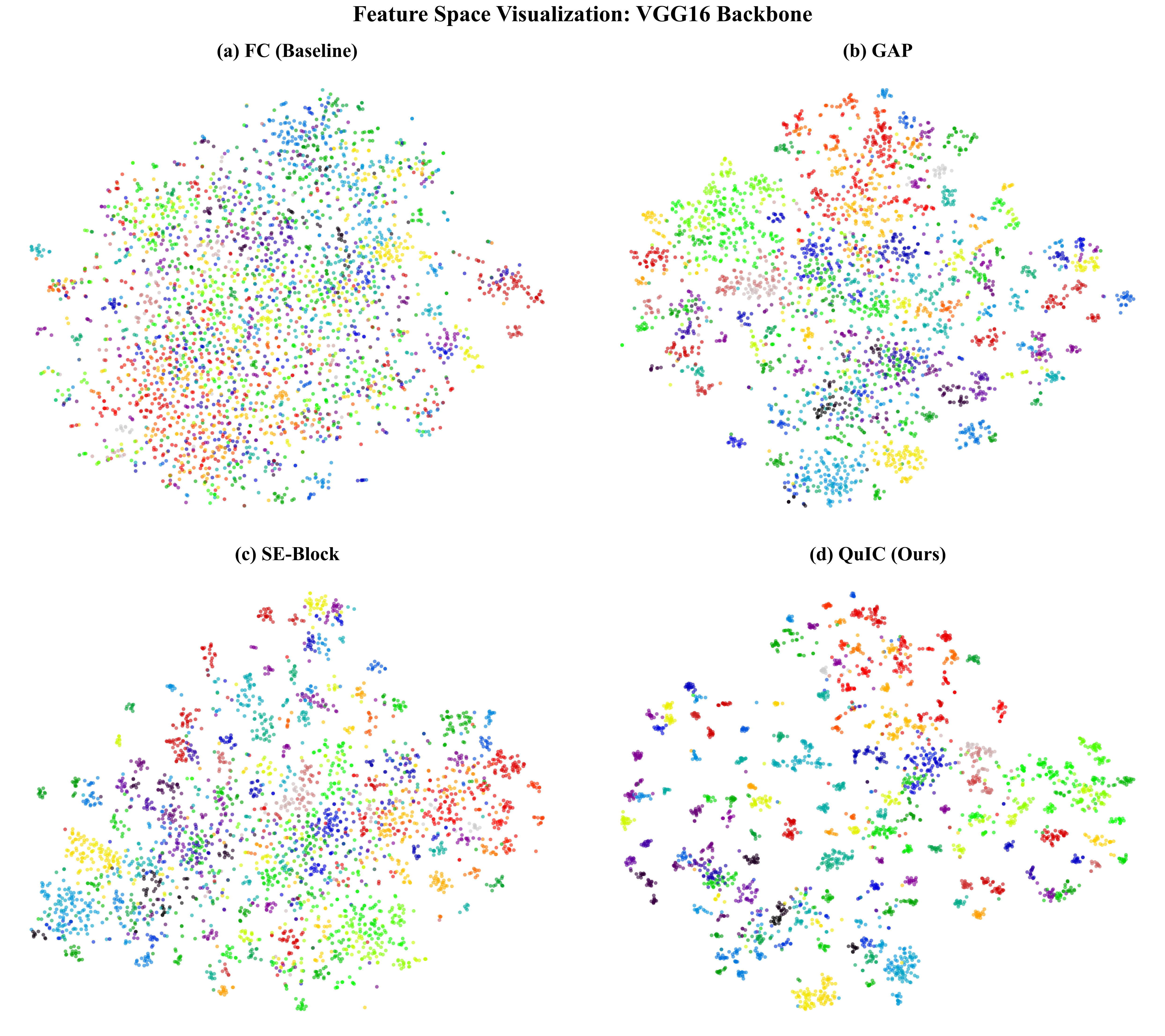}
    \caption{t-SNE visualization of feature spaces on the CUB-200-2011 test set using the VGG16 backbone. Compared to the diffuse, overlapping clusters in (a) FC and (b) GAP, and the ambiguous boundaries in (c) SE-Block, (d) \textbf{QuIC} produces significantly more compact clusters with clear inter-class margins. This demonstrates QuIC's ability to revitalize the shallow backbone by enforcing highly discriminative feature learning.}
    \label{fig:tsne}
\end{figure}

\subsection{Feature Space Visualization}
To further validate the "revitalization" hypothesis, we visualized the feature distributions learned by different classification heads using t-SNE \cite{maaten2008visualizing}. We selected the \textbf{VGG16} backbone as the representative case, as it showed the most significant performance gain (+19.7\%) in our quantitative experiments.

As shown in Fig. \ref{fig:tsne}, the feature manifolds learned by the baselines exhibit distinct topological limitations. The \textbf{FC} and \textbf{GAP} methods (Fig. \ref{fig:tsne} (a) \& (b)) produce diffuse feature clouds with severe overlap between categories, indicating that the legacy backbone struggles to project fine-grained categories into linearly separable regions. While the \textbf{SE-Block} (Fig. \ref{fig:tsne} (c)) improves local clustering by reweighting informative channels, the decision boundaries remain ambiguous with no clear margins.

In stark contrast, \textbf{QuIC} (Fig. \ref{fig:tsne} (d)) transforms the feature space into highly compact, island-like clusters with distinct inter-class margins. This visual evidence confirms that explicitly modeling feature entanglement acts as a strong regularizer, forcing the shallow backbone to maximize both intra-class compactness and inter-class separability.

\section{Conclusion}

In this paper, we presented the \textbf{Quantum-inspired Interaction Classifier (QuIC)}, a novel plug-and-play module designed to revitalize shallow CNNs for Fine-Grained Visual Classification. By mathematically modeling feature vectors as quantum states and the classification head as an observable operator, QuIC effectively captures \textbf{second-order feature interactions} (entanglements) while avoiding the dimensionality explosion associated with traditional bilinear pooling.

Our experiments on the CUB-200-2011 dataset validate that QuIC consistently outperforms standard baselines (GAP and SE-Block). Notably, it delivers a transformative gain to legacy architectures, boosting the Top-1 accuracy of VGG16 by nearly \textbf{20\%}. Qualitative analyses, including Grad-CAM and t-SNE visualizations, confirm that QuIC's success stems from its ability to localize subtle biological features—such as throat hackles and beak shapes—and enforce compact, discriminative feature clustering.

This work suggests that physics-inspired formulations offer a promising, efficient path for designing interpretable deep learning modules that maximize the potential of lightweight backbones.

\bibliographystyle{IEEEtran}
\bibliography{refs}
\end{document}